\documentclass[10pt,twocolumn,letterpaper]{article}

\usepackage{cvpr}              
\usepackage[accsupp]{axessibility} 
\usepackage{graphicx}
\usepackage{amsmath}
\usepackage{amssymb}
\usepackage{booktabs}
\usepackage{multirow}

\usepackage[pagebackref,breaklinks,colorlinks]{hyperref}

\usepackage[capitalize]{cleveref}
\crefname{section}{Sec.}{Secs.}
\Crefname{section}{Section}{Sections}
\Crefname{table}{Table}{Tables}
\crefname{table}{Tab.}{Tabs.}

\def\cvprPaperID{5237} 
\def\confName{CVPR}
\def\confYear{2022}

\begin{document}

\title{PSTR: End-to-End One-Step Person Search With Transformers}

\author{Jiale Cao$^{1,2}$, Yanwei Pang$^{1}\thanks{Corresponding author: Yanwei Pang}$, Rao Muhammad Anwer$^{2}$, Hisham Cholakkal$^{2}$, Jin Xie$^{3,2}$, \\Mubarak Shah$^{4}$, Fahad Shahbaz Khan$^{2,5}$\\
$^1$Tianjin University~~~~~$^2$Mohamed bin Zayed University of Artificial Intelligence \\$^3$Chongqing University~~~~~$^4$University of Central Florida~~~~~ $^5$Linköping University\\
{\tt\small \{connor,pyw\}@tju.edu.cn,~\{rao.anwer,hisham.cholakkal,fahad.khan\}@mbzuai.ac.ae}\\
{\tt\small xiejin@cqu.edu.cn,~shah@crcv.ucf.edu}
}
\maketitle

\begin{abstract}
We propose a novel one-step transformer-based person search framework, PSTR, that jointly performs person detection and re-identification (re-id) in a single architecture. PSTR comprises a person search-specialized (PSS) module that contains a detection encoder-decoder for person detection along with a discriminative re-id decoder for person re-id. The discriminative re-id decoder utilizes a multi-level supervision scheme with a shared decoder for discriminative re-id feature learning and also comprises a part attention block to encode relationship between different parts of a person. We further introduce a simple multi-scale scheme  to support re-id across person instances at different scales. PSTR jointly achieves the diverse objectives of object-level recognition (detection) and instance-level matching (re-id). To the best of our knowledge, we are the first to propose an end-to-end one-step  transformer-based person search framework.
Experiments are performed on two popular benchmarks: CUHK-SYSU and PRW. Our extensive ablations reveal the merits of the proposed contributions. Further, the proposed PSTR sets a new state-of-the-art on both benchmarks. On the challenging PRW benchmark, PSTR achieves a mean average precision (mAP) score of 56.5\%. The source code is available at \url{https://github.com/JialeCao001/PSTR}.
\end{abstract}

\section{Introduction} 
Person search  aims to detect and identify a target person from a gallery of real-world uncropped images, which can be seen as a joint task of person detection \cite{Cao_PedSurvey_TPAMI_2021,Zhang_FasterATT_CVPR_2018,Liu_CSP_CVPR_2019,Pang_MGAN_ICCV_2019}  and re-identification (re-id) \cite{Zheng_ReIDW_CVPR_2017,Liao_LMOR_CVPR_2015,Liao_LMOR_CVPR_2017}. Person search involves addressing the challenges of these two diverse sub-tasks as well as jointly optimizing them in a unified framework.

\begin{figure}[t!]
\centering
\includegraphics[width=0.95\linewidth]{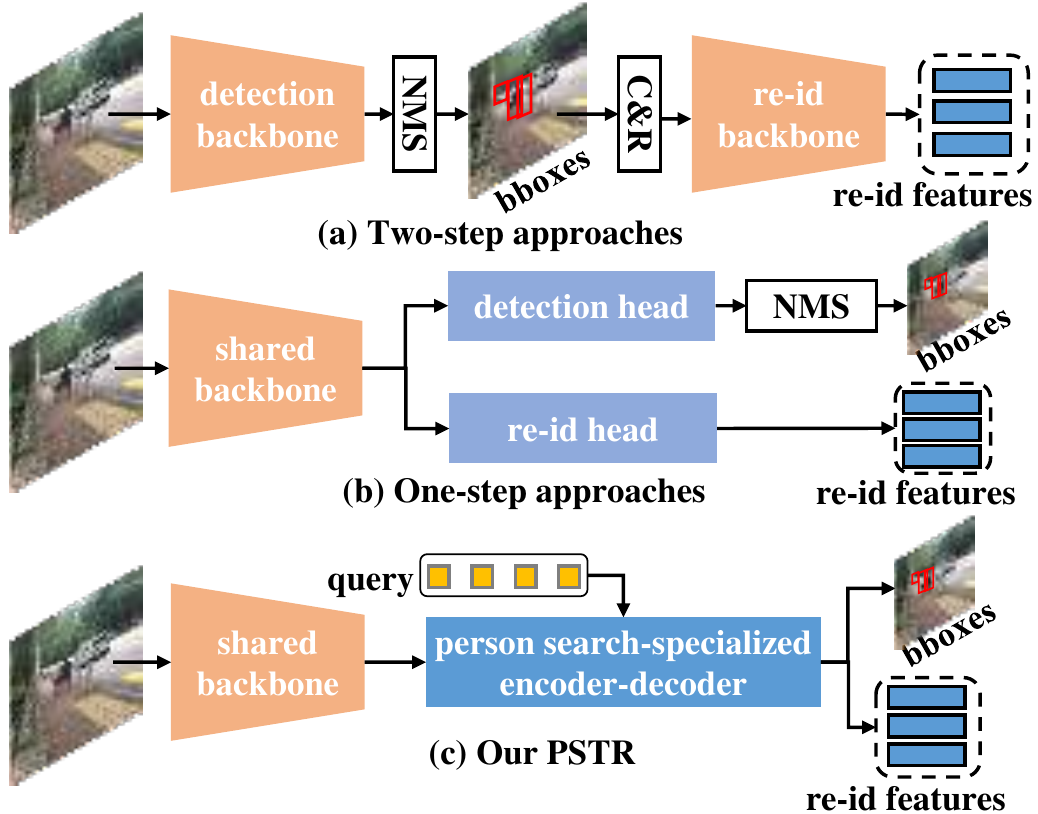}\vspace{-0.3cm}
\caption{Comparison of our PSTR architecture (c) with the existing two-step (a) and one-step paradigms (b). (a) Within the two-step paradigm, person detection and re-id sub-tasks are performed with two separate independent networks. Here, bounding-boxes are first predicted by a detection network and then cropped and resized (C\&R) before being fed to a re-id network. (b) Within the one-step paradigm, detection and re-id branches share the same backbone network.  (c) Distinct from these two paradigms, our PSTR is an end-to-end one-step transformer-based architecture with a person-search specialized module to jointly perform detection and re-id without requiring an NMS post-processing step.}\vspace{-0.4cm}
\label{fig:comp}
\end{figure}

Person search approaches can be roughly divided into two-step \cite{Zheng_PRW_CVPR_2017,Chen_MGTS_ECCV_2018,Han_RDLR_ICCV_2019} and one-step methods\cite{Xiao_OIM_CVPR_2017,Yan_CTXG_CVPR_2019,Chen_NAE_CVPR_2020}.  Two-step approaches 
typically disentangle the two sub-tasks, where person detection and re-id are performed separately (Fig.~\ref{fig:comp}(a)). First, an off-the-shelf detection network (\textit{e.g.,} Faster R-CNN \cite{Ren_FasterRCNN_NIPS_2015}) is employed to detect pedestrians. Second, the detected pedestrians are cropped and resized into a fixed resolution, followed by utilizing a re-id network to identify cropped pedestrians. While achieving promising performance, most two-step approaches are computationally expensive. In contrast, one-step approaches simultaneously detect and identify the persons using a single network (Fig.~\ref{fig:comp}(b)). First, the features are extracted by a shared network. Then, person detection and re-id are performed by two branches within the same network. 

Despite recent progress in person search, both two-step and one-step approaches employ hand-designed mechanisms, such as non-maximum suppression (NMS) procedure to filter out duplicate predictions for each person. Recently, transformers \cite{Vaswani_Att_NIPS_2017,Dosovitskiy_ViT_ICLR_2020} have shown promising results in several vision tasks, including object detection \cite{Carion_DETR_ECCV_2020,Zhu_DeformableDETR_ICLR_2021}. The encoder-decoder design of transformer-based object detectors alleviates the need to employ different hand-designed components, leading to a simpler end-to-end trainable architecture. Further, the transformer architecture can be easily extended to a multi-task learning framework \cite{Kim_HOTR_CVPR_2021,Wang_VISTR_CVPR_2021}. Despite their recent success, transformers are yet to be investigated for person search. In this work, we investigate the problem of designing a simple but accurate end-to-end one-step  transformer-based person search framework.

When designing a one-step transformer-based person search framework, a straight-forward way is to adopt an object detector, such as DETR \cite{Carion_DETR_ECCV_2020} to detect persons, while the re-ID sub-task can be performed in different ways. (i) The transformer decoder within object detector can be modified by introducing an auxiliary task of re-id. (ii) Two  separate standard  encoder-decoder networks can be utilized to perform detection and re-id sub-tasks. However, we observe these strategies struggle to achieve satisfactory results.

\subsection{Motivation}
We consider two desirable properties when designing a transformer-based person search framework.

\noindent\textbf{Improved re-id feature discriminability:} The sub-tasks of detection and re-id within person search have different objectives. Person detection strives to perform \textit{object-level} recognition and localization by differentiating the person category from background. Here, all person instances within and across images are grouped into a single person category. On the other hand, person re-id sub-task aims to identify a person at \textit{instance-level}. Here, a person instance is desired to be matched with a database of images, thereby requiring to discriminate among instances of different persons within the same person category. Therefore, transformer re-id decoders need to be distinct from their detector counterparts and are desired to generate discriminative features specialized to perform instance-level matching. \\
\noindent\textbf{Encoding multi-scale information for re-id:} Scale variation is a challenging problem in person search. The same person captured by different cameras may have a large variation in scale, which increases the difficulty for person matching. Most existing  approaches either follow the strategy where pedestrians are first detected and then resized into a fixed resolution or adopt a feature RoI pooling scheme \cite{Ren_FasterRCNN_NIPS_2015} to obtain scale-invariant representation.  
Instead of image resizing or feature pooling, we look into an  approach to encode multi-scale information within a transformer architecture for re-id in person search. 

\begin{figure}[t!]
\centering
\includegraphics[width=0.95\linewidth]{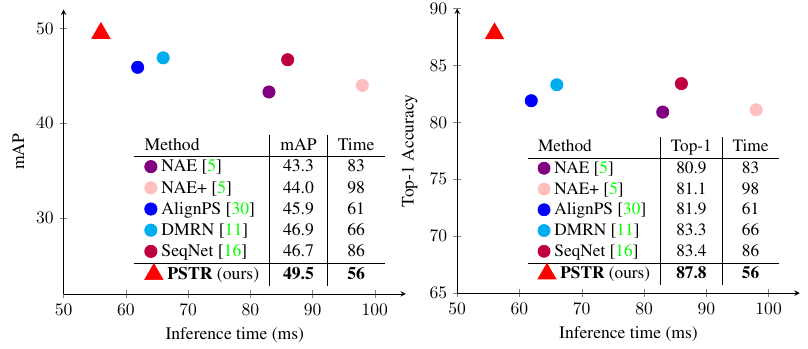}\vspace{-0.3cm}
\caption{Accuracy (AP) vs. speed (ms) comparison with existing one-step methods on PRW test set.  All methods use a ResNet50 backbone and the speed is reported on a V100 GPU.  Our end-to-end one-step transformer-based PSTR outperforms existing methods in terms of both speed \textit{and} accuracy.   } \vspace{-0.2cm}
\label{intro_fig}
\end{figure}

\subsection{Contributions}
We propose a novel end-to-end one-step transformer-based person search framework, named PSTR. Our PSTR treats person search as a sequence prediction problem, where  all persons in an image are  detected along with their respective re-id features (Fig.~\ref{fig:comp}(c)).
To this end, we introduce a person search-specialized (PSS) module  within PSTR that performs both detection and re-id. The PSS module aims to improve  feature discriminability of re-id features by introducing a discriminative re-id decoder that utilizes a multi-level supervision scheme with a shared decoder design. Further, we introduce a part attention block within the discriminative re-id decoder to capture the relationship of different parts. Moreover, we propose a simple multi-scale scheme of our discriminative re-id decoder to address the issue of person matching at different scales. To the best of our knowledge, PSTR is the first end-to-end one-step person search framework based on transformers. 

We validate PSTR on  CUHK-SYSU~\cite{Xiao_OIM_CVPR_2017} and PRW~ \cite{Zheng_PRW_CVPR_2017}.
Our comprehensive ablations reveal the merits of the contributions. Further, PSTR sets a new state-of-the-art on both benchmarks. When using ResNet50~\cite{He_ResNet_CVPR_2016}, PSTR achieves a mAP score of 49.5\% on PRW benchmark, while running at a speed of 56 milliseconds (ms) on a single V100 GPU (see Fig. \ref{intro_fig}). With a transformer-based backbone~\cite{Wang_PVTv2_arXiv_2021}, PSTR obtains the best reported results with a mAP score of 56.5\%. 

\section{Related Work}


\begin{figure*}[t!]
\centering
\includegraphics[width=0.98\linewidth]{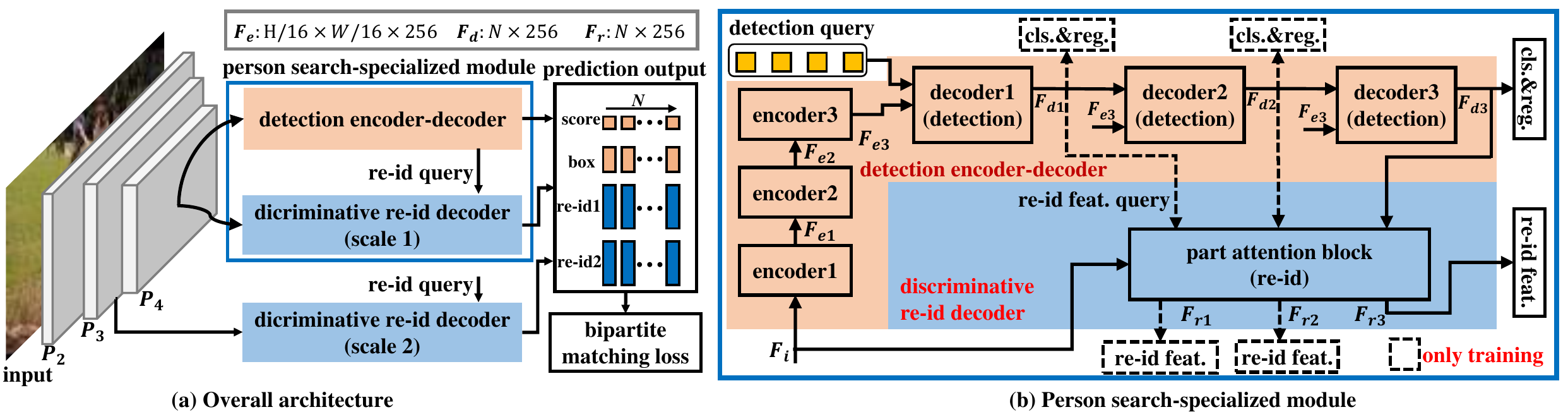} \vspace{-0.3cm}
\caption{(a) Overall architecture of our end-to-end one-step PSTR. PSTR comprises a backbone and a person search-specialized (PSS) module designed to perform detection and re-id for person search. (b) The PSS module consists of a detection encoder-decoder and a novel discriminative re-id decoder. The detection encoder-decoder takes backbone features and performs pedestrian regression and classification using three cascaded decoders followed by a prediction head. The discriminative re-id decoder utilizes a multi-level supervision scheme with a shared decoder 
that takes re-id feature queries from one of the three detection decoders as input during training. The multi-level supervision scheme enables diversity in detected box locations and input re-id feature queries, thereby enhancing the discriminability of re-id features.
We further introduce a part attention block in discriminative re-id decoder to capture the relationship between different parts of a person. The PSS module is utilized in a multi-scale extension  to support re-id across person instances at different scales.} \vspace{-0.4cm} 
\label{fig:arch}
\end{figure*}

\noindent\textbf{Person search:} Person search approaches can be roughly divided into two-step and one-step methods. To address the sub-tasks of detection and re-id, two-step approaches \cite{Zheng_PRW_CVPR_2017,Han_RDLR_ICCV_2019,Lan_CLSA_ECCV_2018} utilize two separate networks dedicated for detection \cite{Ren_FasterRCNN_NIPS_2015,Nie_EGFR_ICCV_2019,Pang_EFIP_CVPR_2019} and re-id \cite{Ye_ReIDSurvey_TPAMI_2020,Cao_SipMask_ECCV_2020}. Zhang \textit{et al.} \cite{Zheng_PRW_CVPR_2017}  explore person search by introducing two independent models. Chen \textit{et al.} \cite{Chen_MGTS_ECCV_2018} propose a mask-guided two-stream network to obtain enhanced feature representation. Wang \textit{et al.} \cite{Cheng_TCTS_CVPR_2020} utilize an identity-guided query detector to extract the query-like proposals and employ a detection adapted model for re-id.

One-step person search methods  integrate detection and re-id into a unified framework. Xiao \textit{et al.} \cite{Xiao_OIM_CVPR_2017} introduce a re-id branch into Fast R-CNN  for person matching. Chen \textit{et al.} \cite{Chen_NAE_CVPR_2020} propose to use norm-aware embedding to separate detection and re-id. Munjal \textit{et al.} \cite{Munjal_QEEPS_CVPR_2019} build the relationship between query image and gallery image by integrating a query-guided Siamese squeeze-and-excitation block into the backbone.  
The work of \cite{Dong_BINet_CVPR_2020} employs a Siamese network that takes input both the entire image and cropped persons  to better guide the feature learning of persons.
Several existing works \cite{Yan_CTXG_CVPR_2019,Chang_RCAA_ECCV_2018,Li_SeqNet_AAAI_2021} explore the problem of utilizing contextual information for person search. Recently, Yan \textit{et al.} \cite{Yan_AlignPS_CVPR_2021} introduce a novel anchor-free approach for person search.

\noindent\textbf{End-to-end object detection with transformers:}
Recently, DETR~\cite{Carion_DETR_ECCV_2020} introduces an end-to-end pipeline for object detection, which predicts objects by a set of detection queries. DETR faces the issues of slow convergence and lower performance on small-sized objects. To solve these issues, deformable DETR~ \cite{Zhu_DeformableDETR_ICLR_2021} replaces standard attention module by a deformable attention module, which focuses on a small set of local sampling points around a reference.  
For an input image, features obtained from the backbone are first enhanced by an encoder. With enhanced features and detection queries, deformable transformer decoder generates $N$ final object features. Finally, a prediction head predicts classification scores and bounding locations.

\section{Method}
\noindent \textbf{Overall architecture:} Fig.~\ref{fig:arch}(a) shows the overall architecture of our PSTR.
We base it  on transformer-based object detector, deformable DETR~\cite{Zhu_DeformableDETR_ICLR_2021}. Our PSTR replaces the standard encoder-decoder in deformable DETR with a person-search specialized (PSS) module (Fig.~\ref{fig:arch}(b)). The PSS module is designed to perform detection and re-id for person search, which comprises a detection encoder-decoder along with a discriminative re-id decoder. The detection encoder-decoder takes backbone features and performs pedestrian regression and classification using three cascaded  decoders followed by a prediction head, as in~\cite{Zhu_DeformableDETR_ICLR_2021}. The discriminative re-id decoder utilizes a multi-level supervision scheme with a shared decoder design by taking re-id feature queries from one of the three detection decoders as input.
It then generates discriminative re-id features for instance-level matching. The multi-level supervision scheme in our discriminative (shared) re-id decoder provides diverse input re-id feature queries and box sampling locations, 
thereby guiding the feature learning for person search. In addition to its shared design, our novel discriminative re-id decoder comprises a part attention decoder to capture the relationship between different person parts. 
To  support re-id across person instances at different scales, we employ our PSS module in a multi-scale extension by using the features of different layers. 
Consequently, the resulting multi-scale re-id features are concatenated to perform instance-level matching with the query person. 
\subsection{Person Search-Specialized Module}
In our PSTR, we obtain  features  from backbone (\eg, ResNet~\cite{He_ResNet_CVPR_2016} or PVT \cite{Wang_PVTv2_arXiv_2021}) and pass it through a deformable convolution layer to extract local information. The resulting  feature  ${\bf P}_i$ is fed to our person search-specialized (PSS) module.
Further, the PSS module takes a set of detection queries as additional inputs,  and generates the features for detection and re-id, respectively.
The PSS module consists of a detection encoder-decoder (Sec.~\ref{sec:PSS_det}) and a discriminative re-id decoder (Sec.~\ref{sec:PSS_disc_reid}). The detection encoder-decoder predicts the features of classification and regression for detection queries. On the other hand, the discriminative re-id decoder extracts re-id features for detection queries. 

\subsubsection{Detection Encoder-Decoder}
\label{sec:PSS_det}
Within the PSS module, the detection encoder-decoder is built on deformable DETR~\cite{Zhu_DeformableDETR_ICLR_2021}. 
As shown in Fig.~\ref{fig:arch}(b), the detection encoder-decoder  consists  of three encoders and three decoders, utilizing the feature ${\bf P}_i$ as input. 
Each encoder has a deformable self-attention layer and a MLP layer.  The output features  of each encoder  are represented as ${\bf F}_{e1},{\bf F}_{e2},{\bf F}_{e3}$.  Consequently, the first  decoder  takes the ${\bf F}_{e3}$ feature  and the $N$ detection queries as  inputs. Each decoder  contains a standard self-attention layer, a deformable cross-attention layer, and a MLP layer.  The output features  from  each decoder  are represented as ${\bf F}_{d1},{\bf F}_{d2},{\bf F}_{d3}$. We use a feature length of 256 for all the three encoders and  decoders. The decoder features are utilized in a prediction head  for box classification and regression, and these features are further used to obtain re-id feature queries for our discriminative re-id decoder presented next.

\subsubsection{Discriminative Re-id Decoder}
\label{sec:PSS_disc_reid}
We introduce our 
discriminative re-id decoder that produces discriminative re-id features for each person.
Fig.~\ref{fig:arch}(b) shows our discriminative re-id decoder. It takes the feature ${\bf P}_i$  as input.  The discriminative re-id decoder utilizes multi-level supervision with a shared decoder design.
To this end, the discriminative re-id decoder utilizes the  features ${\bf F}_{d1},{\bf F}_{d2},{\bf F}_{d3}$ from different detection decoders as  re-id feature queries to improve the diversity of re-id feature queries and the box locations (sampling locations) during training. During inference, we utilize the feature ${\bf F}_{d3}$ as the re-id feature query to obtain the discriminative re-id feature. We further introduce a part attention block that consists of two part attention layers to capture the relationship between different parts of a person. 
Our discriminative re-id decoder directly operates on the feature $\textbf{P}_i$ by taking re-id queries from the detection decoder. We observe this architectural design to be more accurate for the re-id sub-task, than standard encoder-decoder based design. 

\begin{figure}[t!]
\centering
\includegraphics[width=1.0\linewidth]{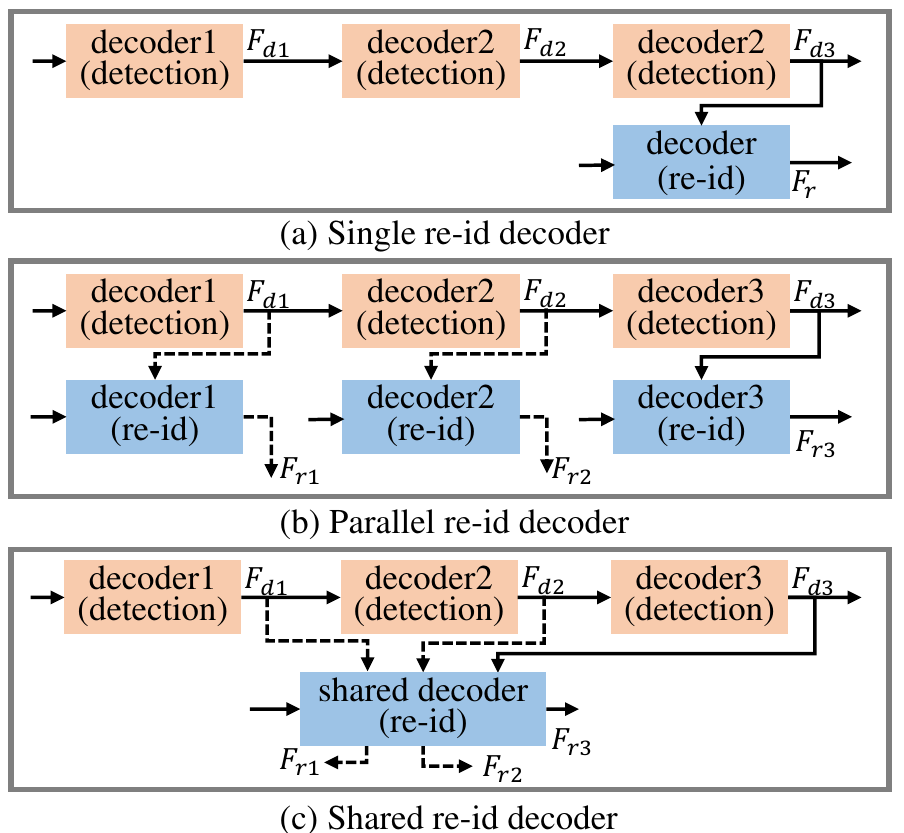} \vspace{-0.6cm}
\caption{Comparison of re-id decoder design schemes. (a) The single-level supervised re-id decoder scheme uses the last detection feature ${\bf F}_{d3}$ as re-id feature query, employing a re-id decoder for feature prediction. Distinct from such a single-level supervised scheme, we introduce two multi-level supervised re-id decoder designs: (b) parallel re-id and (c) shared re-id decoder. (b) The parallel re-id scheme employs three parallel decoder layers to generate re-id features by treating the detection decoder features (${\bf F}_{d1},{\bf F}_{d2},{\bf F}_{d3}$) as queries. Different from the parallel re-id scheme, (c) the shared re-id scheme utilizes a Siamese architecture where all  detection decoders have a common shared re-id decoder to produce corresponding re-id features (${\bf F}_{r1},{\bf F}_{r2},{\bf F}_{r3}$). }\vspace{-0.4cm}
\label{fig:decoder}
\end{figure}

\noindent \textbf{Multi-level supervision with shared decoder design:} A straightforward way to design the re-id decoder is to use the last detection feature ${\bf F}_{d3}$ as the re-id feature query and employ a re-id decoder for feature prediction, as shown in Fig.~\ref{fig:decoder}(a). However, we observe this design to achieve sub-optimal performance likely due to lack of discriminative re-id features learned from a single-level supervision. To this end, we introduce two alliterative schemes that employ multi-level (intermediate) supervision  within the re-id decoder for better re-id feature learning. We call the two proposed schemes as \emph{parallel re-id decoder} and \emph{shared re-id decoder}. Fig. \ref{fig:decoder}(b) and Fig. \ref{fig:decoder}(c) show the two proposed schemes. The parallel re-id decoder treats the  features from each detection decoder as the re-id feature queries and employs three parallel decoder layers to generate the re-id features ${\bf F}_{r1},{\bf F}_{r2},{\bf F}_{r3}$. Here, the re-id features ${\bf F}_{r1},{\bf F}_{r2}$ are only used during training to provide multi-level (intermediate) supervision. Different to the parallel re-id decoder scheme, the shared re-id decoder scheme employs a Siamese architecture where all three re-id feature  queries have 
a shared decoder to generate three re-id features. Similar to parallel re-id decoder, the shared re-id decoder scheme also utilizes the features ${\bf F}_{r1},{\bf F}_{r2}$ only during training.

As discussed earlier, the two sub-tasks of person detection and re-id within person search have diverse  objectives (object-level recognition and instance-level matching). Based on this, we directly utilize the backbone features as input to the 
discriminative re-id decoder, instead of  using the features from the detection encoder. We empirically validate that this leads to superior performance, compared to using features from the detection encoder.  

\begin{figure}[t!]
\centering
\includegraphics[width=1.0\linewidth]{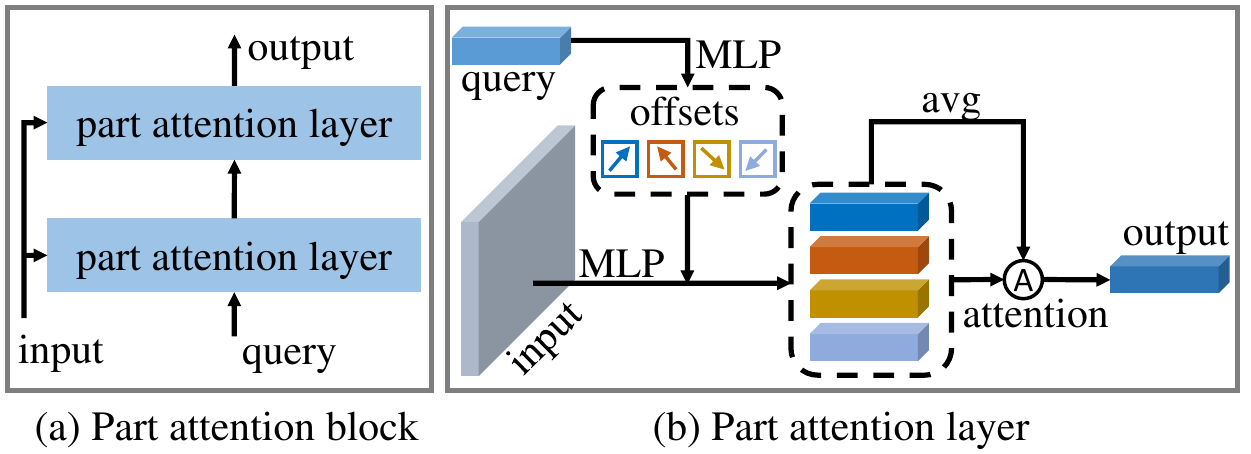}\vspace{-0.2cm}
\caption{(a) The part attention block in our discriminative re-id decoder. The block comprises two part attention layers to encode the relationship between different parts (points).   (b) The part attention layer utilizes  query features to predict the sampling points. The features corresponding to these sampled points are aggregated by fusing the  features from different parts through  cross-attention.}\vspace{-0.4cm}
\label{fig:partatt}
\end{figure}

\noindent \textbf{Part attention block:} To encode the relationship between different parts of a person, we introduce a part attention block (see Fig.~\ref{fig:partatt}) in our discriminative re-id decoder, that employs two layers. Similar to deformable attention~\cite{Zhu_DeformableDETR_ICLR_2021}, we use query features to predict the sampling points, which represent different parts of a person. However, we observe that the attention weights from the query struggles to effectively capture part relations within a person instance. Therefore, different from standard deformable attention, we do not use attention weights from query feature. 
Our part attention averages features at sampling points and then aggregates features from different parts by adapting cross-attention module to generate the output.

\subsection{Multi-Scale Discriminative Re-id Decoder}
Scale variation poses a major challenge in person matching, since the same person can be captured by different cameras at different scales. To address this issue, we introduce a simple extension of our discriminative re-id decoder by employing it at different scales. Here, the discriminative re-id decoders at different scales employ the detection decoder features  as the re-id feature queries. To extract multi-scale re-id features, the additional re-id decoders employ the  features (\textit{e.g.,} $\textbf{P}_2,\textbf{P}_3$) as the input and perform re-id feature generation. During training, these discriminative re-id decoders are supervised by independent re-id losses. During inference, we concatenate these discriminative re-id features from different scales and obtain the  multi-scale re-id feature for person matching.

\subsection{Training and Inference}
Our PSTR predicts classification score, location, and re-id feature for each detection query in an image. The detection features ${\bf F}_{d1},{\bf F}_{d2},{\bf F}_{d3}$ respectively go through two MLP layers for classification and localization. The features ${\bf F}_{r1},{\bf F}_{r2},{\bf F}_{r3}$  are directly used as the re-id features.

During training, we build a lookup table $V$ and a circular queue $U$ to guide re-id feature learning.  We store re-id features of all $L$ labeled   identities in $V$ and  re-id features of $Q$ unlabeled identities from recent mini-batches in $U$. At each iteration, we first compute similarities between re-id features (\textit{e.g.,} $\textbf{F}_{r1}$) in current mini-batch and all features in $V$ and $U$. Then, we compute online instance matching (OIM) loss (described below) based on similarities. During backward propagation, if the re-id feature in mini-batch belongs ground truth identity $i$, we update the $i$-th entry of $V$. We simultaneously push the re-id features of new unlabelled identities into $U$ by popping
older ones. The  OIM loss \cite{Xiao_OIM_CVPR_2017} maximizes expected log-likelihood of each re-id feature in current mini-batch, \textit{i.e.,} ${L}_{\text{oim}} =  \text{log}~p_t$. Here, $p_t$ is the probability of a re-id feature  belonging to the ground truth identity $t$, computed based on the  similarities between a re-id feature and the features at  $V$ and $U$. Finally, the overall loss can be write as 
${L} = \lambda_1 {L}_{\rm cls}+\lambda_2 {L}_{\rm iou}+\lambda_3 {L}_{\rm l1}+\lambda_4 {L}_{\rm oim}$. ${L}_{\rm cls}$ represents  classification loss,  ${ L}_{\rm iou}$ represents bounding-box IoU loss,  ${L}_{\rm l1}$ represents bounding-box $\ell_1$ cost,  and ${L}_{\rm oim}$ represents OIM loss. $\lambda_1,\lambda_2,\lambda_3,\lambda_4$ are the hyper-parameters to balance different losses, which are set as 2.0, 5.0, 2.0, 0.5. 

During inference, we search an annotated (bounding box)  query person in a given query image from  a set of gallery images.
First, we generate multiple predictions of query image using our PSTR, where each prediction includes a classification score, a bounding box and a re-id feature.
Then, the re-id feature of query person is set as the re-id feature of a prediction having maximum overlap with query person bounding box.
Finally, we generate the predictions for all the gallery images, and compute re-id feature similarities  of query person and predictions in gallery images to identify matching persons in gallery images.

\section{Experiments}
\subsection{Datasets and Implementation Details}
\noindent\textbf{CUHK-SYSU} \cite{Xiao_OIM_CVPR_2017} is a large-scale person search dataset.
There are a total 18,184 images covering various real-world challenges, including  viewpoint changes, illumination variations, and diverse backgrounds. It has 96,143 annotated pedestrians, with 8,432 different identities.  
The training set includes 11,206 images, 55,272 pedestrians, and 5,532 identities. The test set contains 6,978 images, 40,871 pedestrians, and 2,900 identities. During inference, the dataset defines a gallery set with different sizes ranging from 50 to 4,000. As in ~\cite{Yan_AlignPS_CVPR_2021,Xiao_OIM_CVPR_2017}, we perform experiments with the standard setting of gallery size 100. Additionally, we analyze the performance when varying the  the gallery size.

\begin{table}[t!]
\renewcommand{\arraystretch}{1.0}
\begin{center}
\resizebox{\linewidth}{!}{
\begin{tabular}{|l|c|cc|cc|}
\hline
\multirow{2}*{Method}       & \multirow{2}*{Backbone}      & \multicolumn{2}{c|}{CUHK-SYSU} & \multicolumn{2}{c|}{PRW}  \\\cline{3-6}
       &       & mAP   & Top-1 & mAP & Top-1  \\
\hline \hline
\multicolumn{6}{|l|}{\textit{Two-step}}\\
IDE \cite{Xiao_OIM_CVPR_2017}  & ResNet50    &  - & - & 20.5 & 48.3 \\ 
MGTS \cite{Chen_MGTS_ECCV_2018}   & VGG16    &  83.0 & 83.7 & 32.6 & 72.1 \\ 
CLSA \cite{Lan_CLSA_ECCV_2018}  & ResNet50    &  87.2 & 88.5 & 38.7 & 65.0 \\ 
RDLR \cite{Han_RDLR_ICCV_2019}  & ResNet50    &  93.0 & 94.2 & 42.9 & 70.2 \\ 
IGPN  \cite{Dong_IGPN_CVPR_2020} & ResNet50    &  90.3 & 91.4 & {47.2} & 87.0 \\ 
TCTS \cite{Cheng_TCTS_CVPR_2020}  & ResNet50    &  {93.9} & {95.1} & 46.8 & {87.5} \\ 
\hline \hline
\multicolumn{6}{|l|}{\textit{One-step with two-stage detector}}  \\ 
OIM \cite{Xiao_OIM_CVPR_2017}      & ResNet50    & 75.5 & 78.7  & 21.3 & 49.4 \\
IAN \cite{Xiao_IAN_PR_2019}      & ResNet50    & 76.3 & 80.1  & 23.0 & 61.9 \\
NPSM \cite{Liu_NPSM_ICCV_2017}   & ResNet50    & 77.9 & 81.2  & 24.2 & 53.1 \\
RCAA \cite{Chang_RCAA_ECCV_2018}   & ResNet50    & 79.3 & 81.3  & - & - \\
CTXG  \cite{Yan_CTXG_CVPR_2019}     & ResNet50    & 84.1 & 86.5  & 33.4 & 73.6 \\
QEEPS \cite{Munjal_QEEPS_CVPR_2019}      & ResNet50    & 88.9 & 89.1  & 37.1 & 76.7 \\
BINet \cite{Dong_BINet_CVPR_2020}      & ResNet50    & 90.0 & 90.7  & 45.3 & 81.7 \\
APNet \cite{Zhong_APNet_CVPR_2020}      & ResNet50    & 88.9 & 89.3  & 41.9 & 81.4 \\
NAE \cite{Chen_NAE_CVPR_2020}      & ResNet50    & 91.5 & 92.4  & 43.3 & 80.9 \\
NAE+ \cite{Chen_NAE_CVPR_2020}       & ResNet50    & 92.1 & 92.9  & 44.0 & 81.1 \\
PGSFL \cite{Kim_PGSFL_CVPR_2021}   & ResNet50    & 90.2 & 91.8  & 42.5 & 83.5 \\
PGSFL \cite{Kim_PGSFL_CVPR_2021}   & ResNet50-dilated   & 92.3 & 94.7  & 44.2 & 85.2 \\
SeqNet \cite{Li_SeqNet_AAAI_2021}      & ResNet50    & 93.8 & 94.6  & 46.7 & 83.4 \\
DMRN    \cite{Han_DMRN_AAAI_2021}   & ResNet50    & 93.2 & 94.2  & 46.9 & 83.3 \\
\hline
\multicolumn{6}{|l|}{\textit{One-step with anchor-free detector}}  \\ 
AlignPS \cite{Yan_AlignPS_CVPR_2021}       & ResNet50    & 93.1 & 93.4  & 45.9 & 81.9 \\
AlignPS \cite{Yan_AlignPS_CVPR_2021}       & ResNet50-DCN    & 94.0 & 94.5  & 46.1 & 82.1 \\
\hline
\multicolumn{6}{|l|}{\textit{One-step with end-to-end transformer}} \\ 
PSTR  (Ours)      & ResNet50   & 93.5 &  95.0  & 49.5 &  87.8 \\
PSTR  (Ours)      & ResNet50-DCN    & 94.2 & 95.2 &50.1 &  87.9   \\
PSTR (Ours)      & PVTv2-B2    & \textbf{95.2} & \textbf{96.2}  & \textbf{56.5} & \textbf{89.7} \\
\hline
\end{tabular}} \vspace{-0.5cm}
\end{center}
\caption{State-of-the-art comparison in terms of mAP and top-1 accuracy on CUHK-SYSU and PRW test sets. When compared with the recently introduced anchor-free AlignPS on CHUK-SYSU, PSTR achieves favorable results in terms of both mAP and top-1 accuracy, using the same ResNet50 backbone. On PRW, PSTR achieves absolute gains of 3.6\% and 5.9\% in terms of mAP and top-1 accuracy, respectively over AlignPS using the same ResNet50 backbone.
Further, PSTR obtains the best reported results using the transformer-based PVT backbone on both datasets.} \vspace{-0.4cm}
\label{tab_stateofart}
\end{table}

\noindent\textbf{PRW} \cite{Zheng_PRW_CVPR_2017} is a challenging person search dataset collected by 6 static cameras.  
The training set contains 5,704 images, 18,048 pedestrians, and 482 identities. The test set has 6,112 images, 25,062 pedestrians, and 450 identities. 

\noindent\textbf{Evaluation metrics:} We employ two standard metrics for person search performance evaluation: mean Averaged Precision (mAP) and top-1 accuracy. \\
\noindent\textbf{Implementation details:} We conduct experiments with two ImageNet \cite{Russakovsky_ImageNet_IJCV_2015} pre-trained backbones: ResNet50 \cite{He_ResNet_CVPR_2016} and recently introduced transformer-based PVTv2-B2 \cite{Wang_PVTv2_arXiv_2021}, which have similar parameters. Our PSTR is trained on a single Tesla V100 GPU using AdamW  optimizer. During training, we employ a multi-scale training scheme and focal OIM loss as AlignPS \cite{Yan_AlignPS_CVPR_2021}. Further, we rescale the test images to a fixed size of  $1500\times900$ pixels  during inference. 
The model is trained for a total of 24 epochs and we use a mini-batch size of 2. The initial learning rate is  set to 0.0001 and we decrease the learning rate by a factor of 10 at  19$^{th}$ and 23$^{th}$ epochs. We will support it with MindSpore.

\subsection{State-of-the-art Comparison}
Here, we compare our one-step transformer-based PSTR with state-of-the-art two-step  and one-step methods.    \\
\noindent \textbf{Comparison on CUHK-SYSU:} Tab.~\ref{tab_stateofart} shows the performance on CUHK-SYSU test set~\cite{Xiao_OIM_CVPR_2017} with the gallery size of 100. Among existing two-step methods IGPN \cite{Dong_IGPN_CVPR_2020} and TCTS \cite{Cheng_TCTS_CVPR_2020} achieve mAP scores of 90.3\% and 93.9\%, respectively. Among one-step with two-stage detection-based methods, SeqNet \cite{Li_SeqNet_AAAI_2021} and DMRN \cite{Han_DMRN_AAAI_2021} obtain mAP scores of 93.8\% and 93.2\%, respectively. The recently introduced one-step anchor-free AlignPS~\cite{Yan_AlignPS_CVPR_2021}  with the same ResNet50 backbone achieves mAP score of 93.1. Our PSTR with the same backbone achieves a mAP score of 93.5\%. In terms of top-1 accuracy, PSTR achieves 95.0\%, corresponding to an absolute of 1.6\% over the recently introduced AlignPS~\cite{Yan_AlignPS_CVPR_2021}, while operating at a slightly faster speed (AlignPS: 61ms vs. PSTR: 56ms) with the same Resnet50 backbone. Further, when using the transformer-based PVTv2-B2 backbone, the proposed PSTR  achieves improved results with mAP and top-1 accuracy of 95.2\% and 96.2\%, respectively. It is worth mentioning that the parameters of both the PVTv2-B2 and ResNet50 backbones are comparable.

\begin{figure}[t!]
\centering
\includegraphics[width=1.0\linewidth]{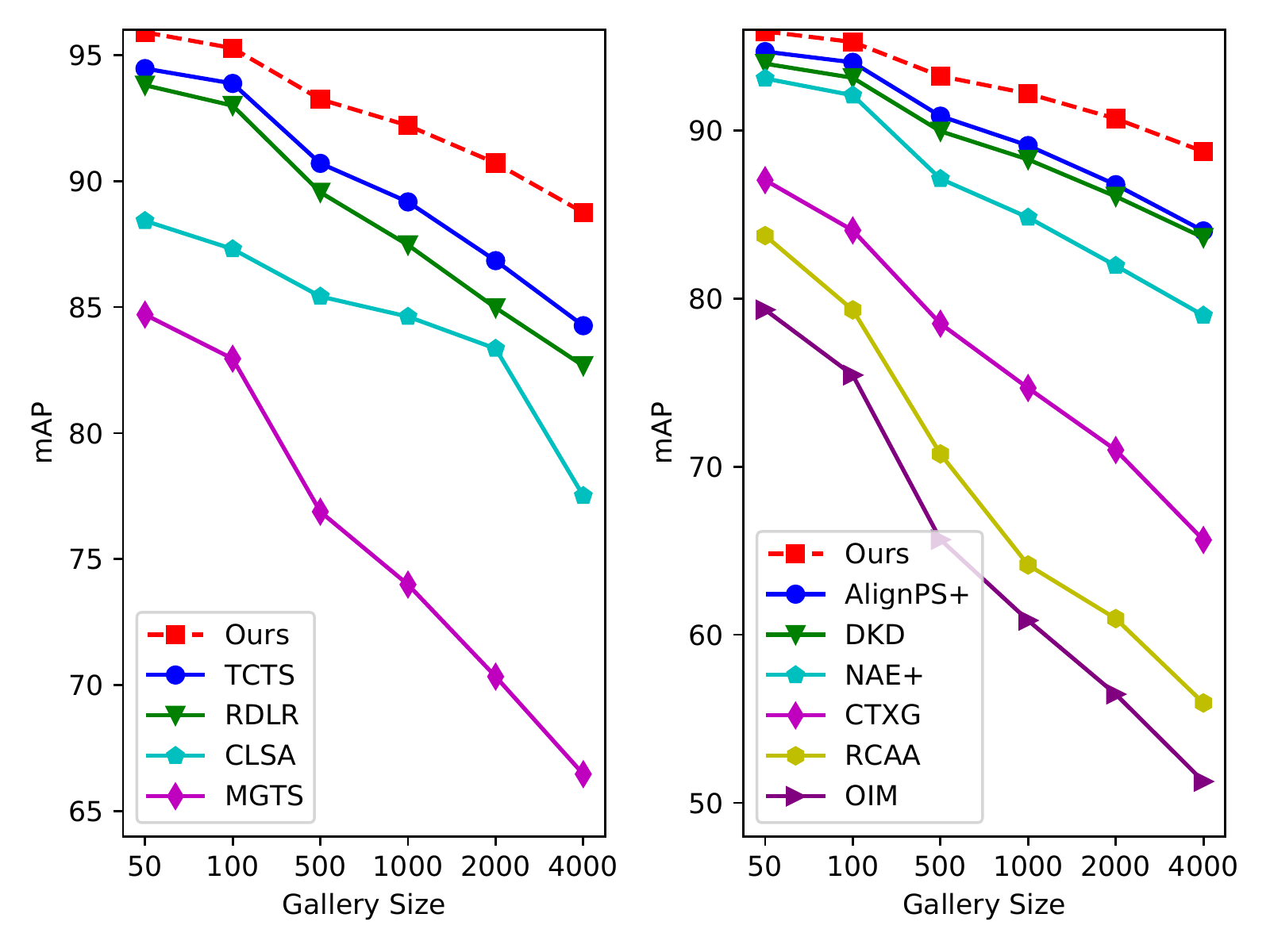}\vspace{-0.4cm}
\caption{State-of-the-art comparison with existing two-step  (left) and one-step methods (right) on CUHK-SYSU dataset~\cite{Xiao_OIM_CVPR_2017} with different gallery sizes. Our PSTR achieves consistent improvement in performance compared to existing methods with different gallery sizes. Further, PSTR outperforms the best existing two-step and one-step methods with a larger performance margin on the more challenging scenario of large gallery size. }\vspace{-0.4cm}
\label{fig:cuhk}
\end{figure}

We further perform a state-of-the-art performance comparison on CUHK-SYSU test set with a gallery size ranging from 50 to 4,000. Fig.~\ref{fig:cuhk} compares our PSTR with existing two-step and one-step approaches in terms of mAP.  Our PSTR consistently outperforms existing person search approaches under different gallery sizes. 

\noindent\textbf{Comparison on PRW:} Tab.~\ref{tab_stateofart} shows the state-of-the-art comparison on the PRW test set~\cite{Zheng_PRW_CVPR_2017}. Among existing two-step approaches, TCTS \cite{Cheng_TCTS_CVPR_2020} and IGPN \cite{Dong_IGPN_CVPR_2020} achieve respective mAP scores of 46.8\% and 47.2\% and top-1 accuracy scores of 87.5\% and 87.0\%. In case of one-step with two-stage detection-based person search methods, DMRN \cite{Han_DMRN_AAAI_2021} and SeqNet \cite{Li_SeqNet_AAAI_2021} obtain respective mAP scores of 46.9\% and 46.7\% and top-1 accuracy scores of 83.3\% and 83.4\%. Further, AlignPS \cite{Yan_AlignPS_CVPR_2021}   achieves mAP and top-1 accuracy of 45.9\% and 81.9\%, respectively. Our PSTR achieves an absolute gain of 3.6 and 5.9 in terms of mAP and top-1 accuracy over AlignPS, using the same ResNet50 backbone. Note that distinct from AlignPS, our transformer-based one-step PSTR does not rely on an NMS post-processing step. 

In addition, we compare with the SeqNet variant (SeqNet$^\dag$) that further introduces a post-processing strategy within SeqNet   to refine the matching scores between query and gallery. For a fair comparison with SeqNet$^\dag$, we introduce same matching scoring strategy in our PSTR (named PSTR$^\dag$) achieving mAP and top-1 accuracy of 50.1\% and 89.2\%, corresponding to absolute gains of 2.5\% and 1.6\% over SeqNet$^\dag$. We also compare with the recent work of \cite{Zhang_DKD_AAAI_2021} that introduces a person search approach, DKD, where a novel knowledge distillation strategy is employed. However, DKD requires training a re-id model and a person search model separately along with careful augmentation design. Further, the DKD requires a longer training (more than 4$\times$ longer compared to ours). Our PSTR achieves favorable top-1 performance (DKD: 87.1\% vs. Ours: 87.8\%), while being 25\% faster at inference. Further, our PSTR does not require a specialized training scheme (separate model for re-id as well as person search and longer training). Lastly,    
we utilize our one-step PSTR (without any post-processing step) with the PVTv2-B2 backbone (Tab.~\ref{tab_stateofart}), achieving the best results reported on this dataset with mAP and top-1 accuracy of 56.5\% and 89.7\%, respectively.

\begin{table}[t!]
\footnotesize
\renewcommand{\arraystretch}{1.0}
\begin{center}
\begin{tabular}{|l|l|cc|}
\hline
\multicolumn{2}{|c|}{Re-id decoder design}      & mAP & Top-1  \\
\hline
\hline
\multirow{3}*{(a) Decoder structure}& Single re-id decoder     & 47.4 & 84.9 \\
& Parallel re-id decoder     & 51.4 & 87.0 \\
& Shared re-id decoder     & 51.9 & 88.3 \\
\hline
\hline
\multirow{3}*{(b) Input feature}& Encoder layer 1     & 49.8 & 88.0 \\
& Encoder layer 2     & 49.3 & 87.4 \\
& Encoder layer 3      & 47.5 & 86.3 \\
& The feature $\textbf{P}_4$     & 51.9 & 88.3 \\
\hline
\hline
\multirow{2}*{(c) Attention layer}& Deformable attention     & 51.5 & 87.8 \\
& Part attention     & 51.9 & 88.3 \\
\hline
\end{tabular}\vspace{-0.5cm}
\end{center}
\caption{Impact of different  design choices in our discriminative re-id decoder. (a) Impact of different decoder structures, shown in Fig. \ref{fig:decoder}, including single re-id decoder, parallel re-id decoder, and shared re-id decoder. (b) Impact of different features for re-id decoder, including the features from detection encoder layers $1,2,3$ and backbone. (c) Impact of different attention layers, including deformable attention and our part attention.} \vspace{-0.4cm}
\label{tab:decoderdesign}
\end{table}

\begin{table}[t!]
\footnotesize
\renewcommand{\arraystretch}{1.0}
\begin{center}
\begin{tabular}{|ccc|cc|}
\hline
Scale 1 ($\textbf{P}_4$)      & Scale 2  ($\textbf{P}_3$)     & Scale 3  ($\textbf{P}_2$)     & mAP & Top-1  \\
\hline \hline
\checkmark       &      &   & 51.9 & 88.3 \\
\checkmark        & \checkmark       &   & 54.7 & 89.0 \\
\checkmark      & \checkmark      &  \checkmark   & 56.5 & 89.7 \\
\hline
\hline
\checkmark       &      &    & 51.7 & 88.0 \\
     & \checkmark      &    & 53.9 & 88.2 \\
     &      &  \checkmark   & 48.5 & 86.7 \\
\hline
\end{tabular}\vspace{-0.5cm}
\end{center}
\caption{Impact of multi-scale re-id decoder. The top part shows the performance of single-scale, two-scale, and three-scale re-id decoders. The bottom part shows performance of individual scales  in our multi-scale re-id decoder with three scales.}\vspace{-0.4cm}
\label{tab:multiscale}
\end{table}

\begin{figure*}[t!]
\centering
\includegraphics[width=0.99\linewidth]{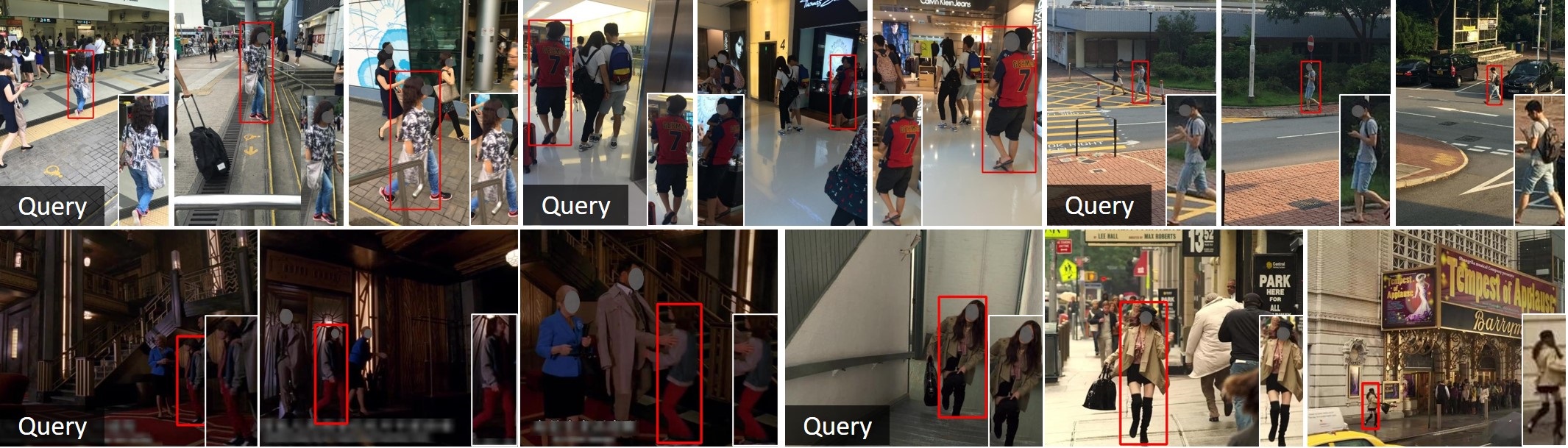}\vspace{-0.3cm}
\caption{Qualitative results on CUHK-SYSU test set \cite{Xiao_OIM_CVPR_2017}. We show the top-two matching results for five different queries. Our PSTR accurately detects and recognizes the query persons under  challenging outdoor and indoor scenes.}\vspace{-0.1cm}
\label{fig:vis-cuhk3}
\end{figure*}

\begin{figure*}[t!]
\centering
\includegraphics[width=0.99\linewidth]{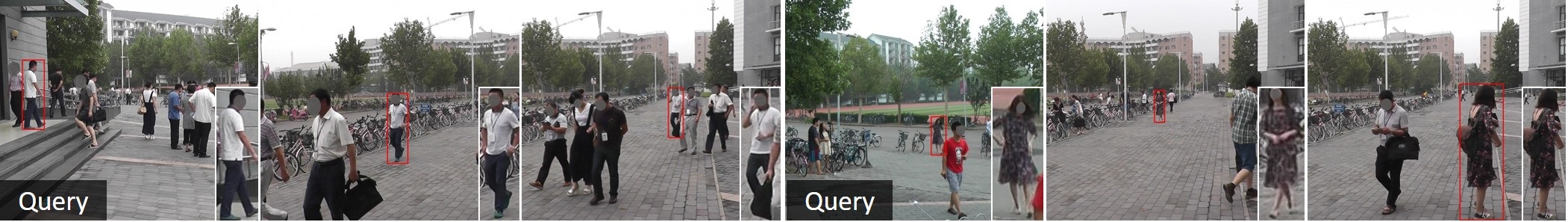}\vspace{-0.3cm}
\caption{Qualitative results on PRW test set \cite{Zheng_PRW_CVPR_2017}. We show the top-two matching results for four different queries. Our PSTR accurately detects and recognizes the query persons across different cameras.}\vspace{-0.4cm}
\label{fig:vis-prw}
\end{figure*} 

\subsection{Ablation Study}
We perform extensive ablations to validate the effectiveness of proposed contributions on PRW. Throughout the ablations, we use same PVTv2-B2 backbone. For a fair comparison, all ablations, except the impact of multi-scale re-id decoder, are performed using re-id decoder at a single scale.\\
\noindent\textbf{Design choices for transformer-based person search:} 
We compare our PSS module with two straightforward strategies for detection and re-id within a one-step transformer pipeline. The first strategy shares the standard encoder-decoder for detection and re-id, achieving 23.1\% mAP and 66.8\% on top-1 accuracy. The second strategy adopts two separate encoder-decoder for detection and re-id, and obtains 44.5\% mAP and 84.6\% top-1 accuracy. Our PSTR with the proposed PSS module achieves superior results with mAP and top-1 accuracy of 51.9\% and 88.3\%, respectively, compared to both these strategies. 

\noindent \textbf{Multi-level supervision with a shared re-id decoder:} Tab.~\ref{tab:decoderdesign}(a) shows the impact of multi-level supervision with a shared decoder design (\textit{shared re-id decoder}) within our re-id decoder, as illustrated Fig.~\ref{fig:decoder}(c). When employing a \textit{single re-id decoder} (Fig.~\ref{fig:decoder} (a)) and taking the features of last detection decoder as input, we obtain 47.4\% on mAP and 84.9\% on top-1 accuracy. Compared to the single-level supervision design, our multi-level supervision scheme (Fig.~\ref{fig:decoder} (b)) in a \textit{parallel re-id decoder} design has the absolute gains of 4.0\% and 2.1\% in terms of mAP and top-1 accuracy, respectively. The best results are obtained when utilizing our multi-level supervision with a shared decoder design (\textit{shared re-id decoder}). It is worth mentioning that all the aforementioned levels of supervisions (single or multi-level) are only utilized during training. At inference, the features from the last detection decoder are used as input to the re-id decoder for all aforementioned methods (a-c). These results show that multi-level supervision with a shared decoder design enables better re-id feature learning.

\begin{figure}[t!]
\centering
\includegraphics[width=\linewidth]{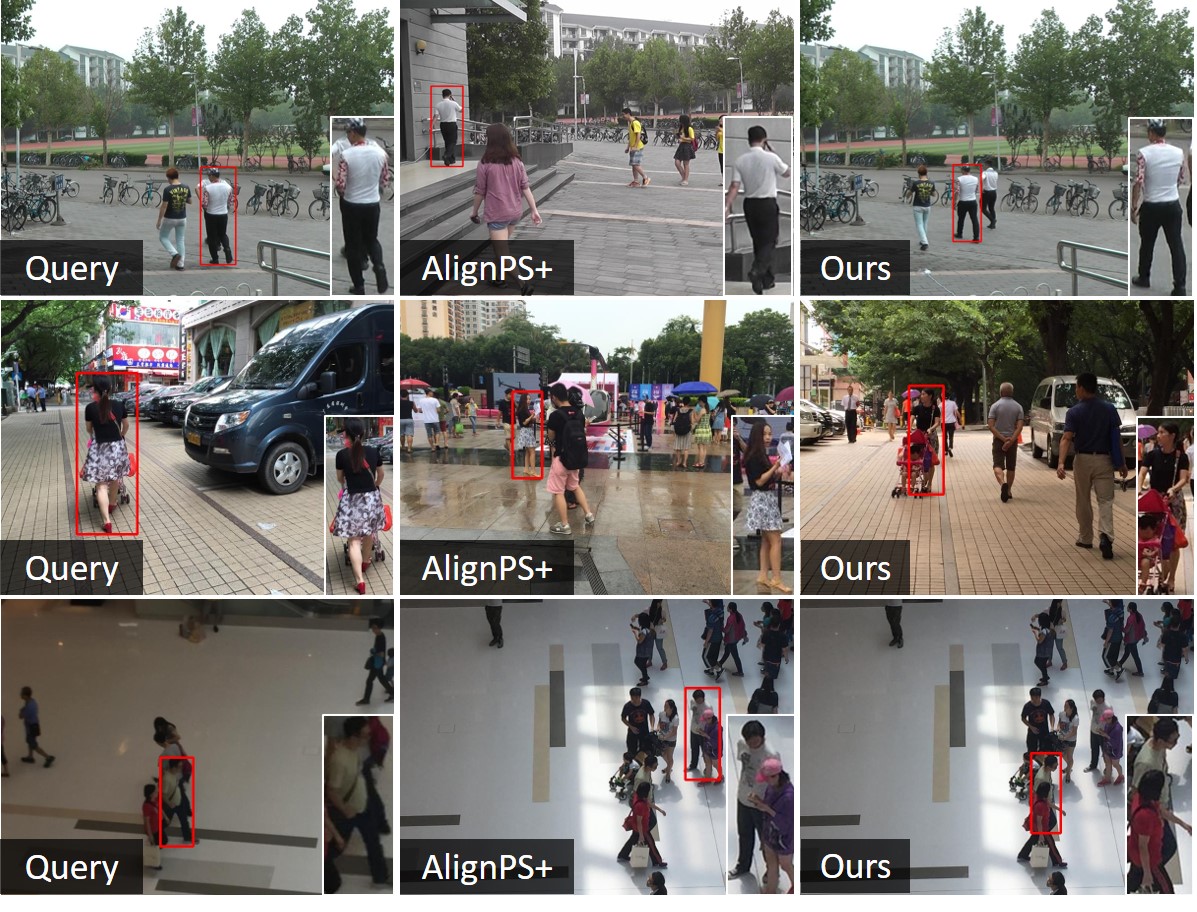}\vspace{-0.4cm}
\caption{Qualitative comparison with AlignPS+~\cite{Yan_AlignPS_CVPR_2021}. We show the top-1 matching results of AligPS+ and our PSTR. For all the  three queries,  our PSTR achieve the correct   matching.}\vspace{-0.4cm}
\label{fig:vis-alignps}
\end{figure}


\noindent \textbf{Impact of input features to re-id decoder:} Tab.~\ref{tab:decoderdesign}(b) shows the impact of different input features on  discriminative re-id decoder.
When utilizing the output features of detection encoder as input features for re-id decoder, there is a drop in performance. 
For example, when taking the output features of last detection encoder as the input of re-id decoder, the performance drops by 4.4\% in mAP. This drop is likely due to the diverse objectives of detection (object-level recognition) and re-id (instance-level matching).\\
\noindent \textbf{Impact of part attention block:}
We also compare the standard deformable attention with our part attention in Tab.~\ref{tab:decoderdesign}(c). Part attention provides 0.4\% improvement on mAP and 0.5\% improvement on top-1 accuracy.

\noindent \textbf{Multi-scale re-id decoder:}  Lastly, we analyze the impact of multi-scale re-id decoder by comparing it with the single-scale, two-scale and three-scale variant in Tab. \ref{tab:multiscale}. We show impact of multi-scale re-id decoder, including single-scale re-id decoder, two-scale re-id decoder, and three-scale re-id decoder. Multi-scale re-id decoder takes the last $N$ feature maps as the inputs for different re-id decoder branches. Three-scale re-id decoder achieves the best performance. 
Further, we show the performance of individual scales in  our three-scale re-id decoder.

\noindent \textbf{Qualitative results:}
We first provide some qualitative comparisons between our PSTR with state-of-the-art AlignPS+ \cite{Yan_AlignPS_CVPR_2021} in Fig. \ref{fig:vis-alignps}. For a given query person, the top-1 matching result is shown. Compared to AlignPS+, our PSTR successfully detects and recognizes the persons in different scenes. We further show some qualitative results on CUHK-SYSU test set \cite{Xiao_OIM_CVPR_2017} and PRW test set \cite{Zheng_PRW_CVPR_2017} in Fig. \ref{fig:vis-cuhk3} and Fig. \ref{fig:vis-prw}.  Our PSTR accurately identifies the query person in the gallery images under different challenging scenes. 

\section{Conclusion and Limitations}
We proposed an end-to-end one-step transformer-based person search approach, named PSTR. Within PSTR, we introduced a novel 
person search-specialized (PSS) module for detection and re-id.
The PSS module comprises a detection encoder-decoder  and a discriminative re-id decoder that employs a multi-level supervision scheme with a shared decoder for better re-id feature learning. Further, it utilizes a part attention block to capture relationship between different parts. Moreover, we introduce a simple multi-scale extension of our re-id decoder. Experiments on two benchmarks reveal benefits of the proposed contributions, leading to state-of-the-art results on both datasets. We observe our PSTR to occasionally struggle at heavy occlusion or extreme low-light conditions. We will be exploit it in future.

Similar to other vision tasks (\textit{i.e.,} face recognition), person search may invade personal privacy if deployed
irresponsibly. It is important to establish relevant laws and policies to protect the privacy when using person search
or other vision technologies for the security of citizens in future. 

\footnotesize{\textbf{Acknowledgement}: This work was supported by National Key R\&D Program (2018AAA0102800), National Natural Science
Foundation of China (61906131),  Tianjin Natural Science
Foundation (21JCQNJC00420), and CAAI-Huawei MindSpore Open Fund.}

{\small
\bibliographystyle{ieee_fullname}
\bibliography{egbib}
}

\end{document}